\documentclass[11pt]{article}

\usepackage{acl}

\usepackage{times}
\usepackage{latexsym}
\usepackage{multirow} 

\usepackage{stfloats}
\usepackage{placeins}

\usepackage[T1]{fontenc}

\usepackage[utf8]{inputenc}

\usepackage{microtype}

\usepackage{inconsolata}

\usepackage{graphicx}
\usepackage{amsmath}

\usepackage{wrapfig}

\usepackage{enumitem}

\usepackage{hyperref}

\usepackage[ruled,vlined,linesnumbered]{algorithm2e}

\SetKwInput{KwInput}{Input}
\SetKwInput{KwOutput}{Output}
\SetKwInput{KwInit}{Initialize}
\SetKwInput{KwStage}{Stage}
\SetKwComment{Comment}{$\triangleright$\ }{}
\DontPrintSemicolon
\SetAlgoNlRelativeSize{-1}
\SetAlgoInsideSkip{smallskip}
\SetAlgoCaptionSeparator{.}

\usepackage[table]{xcolor}
\definecolor{TableHeaderGray}{HTML}{E6E6E6} 
\definecolor{OurMethodBlue}{HTML}{E8F0FE}   
\definecolor{ZebraGray}{HTML}{F9F9F9}       

\definecolor{StdGray}{gray}{0.45}
\newcommand{\std}[1]{{\color{StdGray}\scriptsize$\pm$ #1}}
\newcommand{\best}[2]{\textbf{#1} \std{#2}}
\newcommand{\res}[2]{#1 \std{#2}}

\usepackage{subcaption}
\usepackage{graphicx}
\usepackage{booktabs}
\usepackage{amsfonts}

\usepackage{amsmath,amssymb,bm}

\newcommand{\R}{\mathbb{R}}
\newcommand{\E}{\mathbb{E}}

\newcommand{\KL}{D_{\mathrm{KL}}}

\newcommand{\Ladv}{\mathcal{L}_{\mathrm{adv}}}
\newcommand{\Lfm}{\mathcal{L}_{\mathrm{fm}}}
\newcommand{\Lrecon}{\mathcal{L}_{\mathrm{recon}}}
\newcommand{\Ldur}{\mathcal{L}_{\mathrm{dur}}}
\newcommand{\Lkl}{\mathcal{L}_{\mathrm{kl}}}
\newcommand{\LFzero}{\mathcal{L}_{F_0}}

\title{NüshuVoice: Reviving the Voice of Endangered Nüshu with Pitch-Aware Text-to-Speech}

\author{
 \textbf{Hongkun Yang\textsuperscript{1}}\thanks{Hongkun Yang, Xinhui Yi, Xiyan Zhao, and Yibo Meng contributed equally.}\thanks{Hongkun Yang and Xinhui Yi are core contributors. The relative order of Hongkun Yang and Xinhui Yi was determined by random draw.},
 \textbf{Xinhui Yi\textsuperscript{2}}\footnotemark[1]\footnotemark[2],
 \textbf{Xiyan Zhao\textsuperscript{2}}\footnotemark[1],
 \textbf{Yibo Meng\textsuperscript{3}}\footnotemark[1],
\\
\textbf{Lionel Z. Wang\textsuperscript{2}}\thanks{Corresponding authors.},
 \textbf{Lixu Wang\textsuperscript{4}},
 \textbf{Yaqi Zhang\textsuperscript{5}},
 \textbf{Ruiqi Chen\textsuperscript{6}},
\\
 \textbf{Xuanyue Zhao\textsuperscript{4}},
 \textbf{Lanxin Zhang\textsuperscript{4}},
 \textbf{Yu Zeng\textsuperscript{7}},
 \textbf{Weijia Chu\textsuperscript{2}},
\\
 \textbf{Yiming Ma\textsuperscript{8}},
 \textbf{Chenyu Liu\textsuperscript{2}},
 \textbf{Jianghao Lin\textsuperscript{7}},
 \textbf{Xin Xu\textsuperscript{2}}\footnotemark[3]
\\
 \textsuperscript{1}Ocean University of China,
 \textsuperscript{2}The Hong Kong Polytechnic University,
 \textsuperscript{3}Cornell University,
\\
 \textsuperscript{4}Nanyang Technological University,
 \textsuperscript{5}Shanghai Jiao Tong University,
 \textsuperscript{6}University of Michigan--Ann Arbor,
\\
 \textsuperscript{7}University of Science and Technology of China,
 \textsuperscript{8}Harbin Institute of Technology
}

\begin{document}
\maketitle
\begin{abstract}
Nüshu is an endangered phonetic script historically used by women in Jiangyong County, southern Hunan, China. While existing computational studies of Nüshu mainly focus on textual digitization and visual recognition, the acoustic reconstruction of its authentic pronunciation remains largely unexplored. Building a Nüshu text-to-speech (TTS) system is particularly challenging because available recordings are extremely limited and mostly consist of isolated syllable-level pronunciations rather than natural sentence-level utterances.  In this work, we introduce \textbf{NüshuVoice}, the first TTS benchmark for Nüshu. We construct a sentence-level Nüshu text-to-audio dataset that aligns standardized Unicode Nüshu text, phonetic transcriptions, standard Chinese translations, and archival recordings. To synthesize speech under this extreme low-resource setting, we propose \textbf{Nüshu-PitchVITS}, an $F_0$-conditioned VITS framework that leverages Nüshu's five-level pitch notation as an explicit prosodic inductive bias. Experimental results show that Nüshu-PitchVITS outperforms strong TTS baselines in spectral fidelity, pitch reconstruction, and human-rated intelligibility. We publicly release the dataset and code at: \url{https://anonymous.4open.science/r/Nvshu-TTS-2EB6}.
\end{abstract}

\section{Introduction}

Nüshu is an endangered script historically used by women in Jiangyong County, southern Hunan, China \cite{zuo2024jiangyong}. It is often regarded as the world’s only known script created and used primarily by women. Its emergence was shaped by the unequal access to formal education and literacy that women faced in premodern China. In this context, Nüshu represents not only a writing system, but also a form of women’s cultural agency under gendered conditions of exclusion. Unlike standard Chinese writing, which is largely morphosyllabic and logographic, Nüshu functions as a phonetic syllabic script. Each character corresponds to a specific syllable in the local Jiangyong dialect, including its initial, final, and tone \cite{congrong2024history}. This property makes Nüshu not merely a written cultural artifact, but also an acoustic heritage. Its preservation therefore depends in part on recovering historically grounded pronunciations.

\begin{figure}
    \centering
    \large
    \includegraphics[width=\linewidth]{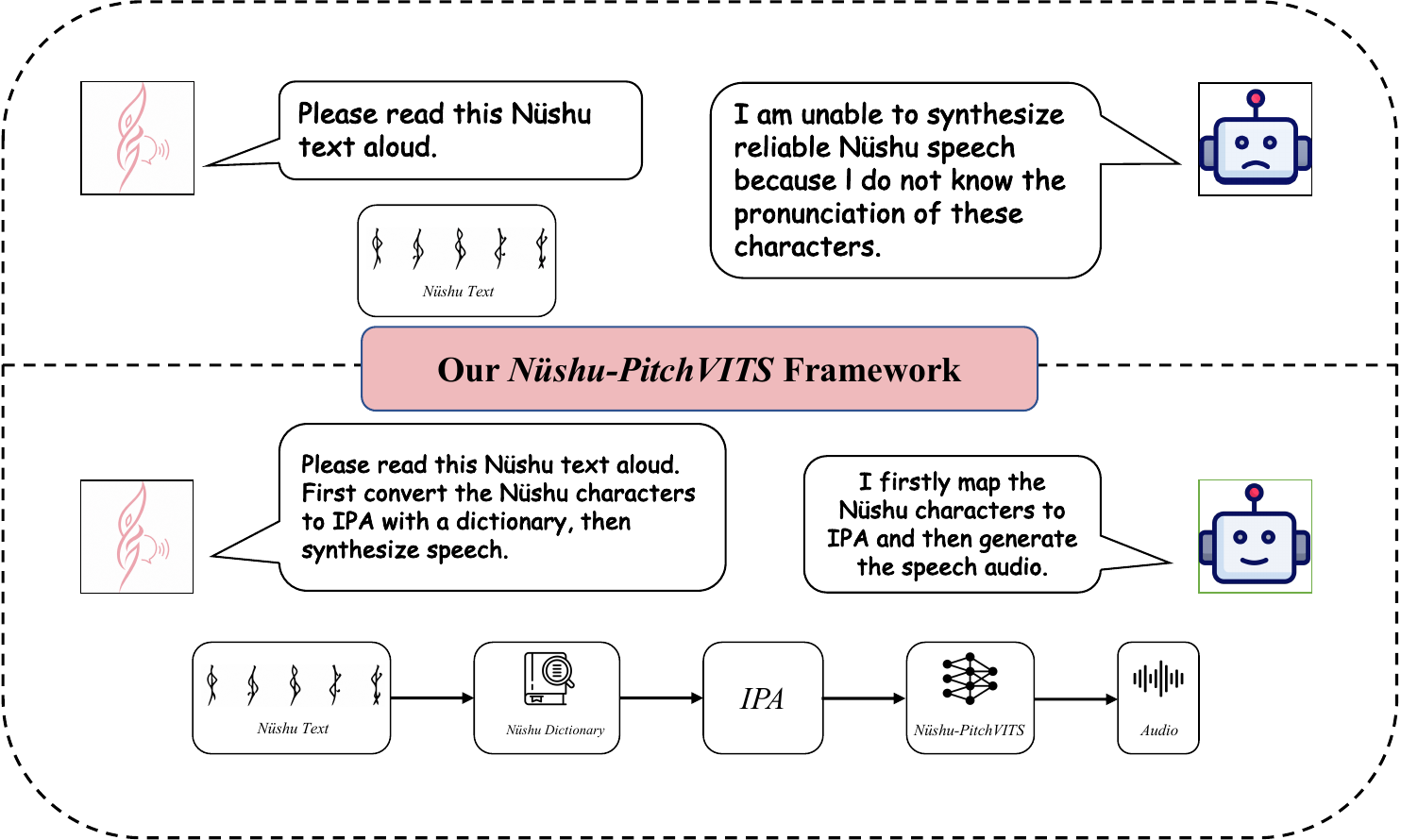}
    \caption{A simplified, stylized illustration of our \textbf{Nüshu-PitchVITS}. Nüshu characters are firstly converted to IPA with a dictionary, and then a TTS model generates speech audio}
    \label{fig:overview}
\end{figure}

Recent work on endangered and Indigenous languages has shown that speech technologies, especially text-to-speech (TTS), can support pronunciation access, language education, and revitalization \cite{chan2025evaluating, hiovain2025world}. For Nüshu, such acoustic access is particularly important because the script is phonetic. A TTS system can therefore serve not only as a tool for generating speech, but also as a computational interface between Nüshu glyphs, phonological annotations, and historically grounded pronunciations. However, existing computational studies of Nüshu have primarily focused on textual digitization, encoding, and visual recognition \cite{sun2023ai,yang2024nushurescue,yang2025recontextualizing}. These efforts are essential for preserving the written form of Nüshu, but they leave its acoustic modality largely underexplored. As a result, Nüshu can be digitally displayed and recognized, yet its spoken realization, tonal dynamics, and dialectal characteristics remain difficult to access for learners, researchers, and cultural heritage applications.

Building a neural TTS system for Nüshu is technically challenging under such preservation-oriented conditions. Modern TTS models typically rely on large amounts of paired text--speech data \cite{xu2020lrspeech, park2023unsupervised, basher2025bntts}, whereas available Nüshu recordings are extremely limited, archival, and non-studio in quality \footnote{Xie, Zhimin, and Wang, Lihua. \textit{Nüshu Fasheng Dianzi Zidian} [\textit{Nüshu Pronunciation Electronic Dictionary}]. Wuhan: Huazhong University of Science and Technology Press, 2002. In Chinese.}. Under this extreme data scarcity, conventional TTS architectures are prone to unstable text-speech alignment and weak prosodic modeling.

In this work, we present the first dedicated study on Nüshu TTS synthesis. We first construct a sentence-level Nüshu TTS dataset by aligning verified Unicode Nüshu text, IPA-based phonetic transcriptions, standard Chinese translations, and sentence-level recordings assembled from archival syllable-level audio. Based on this dataset, we propose an $F0$-conditioned VITS framework tailored to the phonological structure of Nüshu. The key motivation is that Nüshu phonetic annotations use a five-level pitch notation system, which provides explicit tonal information. As shown in \textbf{Figure~\ref{fig:overview}}, Nüshu characters are first mapped into phonetic representations based on the International Phonetic Alphabet (IPA) through a dictionary, and the resulting phonetic sequence is then used by the TTS model to generate speech audio.

Experimental results demonstrate that explicit pitch conditioning substantially improves Nüshu speech synthesis under extreme low-resource conditions. Compared with state-of-the-art baselines, our method achieves a 40.1\% reduction in MCD, a 43.5\% lower $F_0$ RMSE, and a 27.4\% gain in intelligibility MOS over the strongest baseline.. These results suggest that linguistically grounded prosodic modeling is critical for endangered phonetic scripts where acoustic data are scarce but tonal annotations are available.

\begin{figure*}[t]
    \centering
    \includegraphics[width=0.98\linewidth]{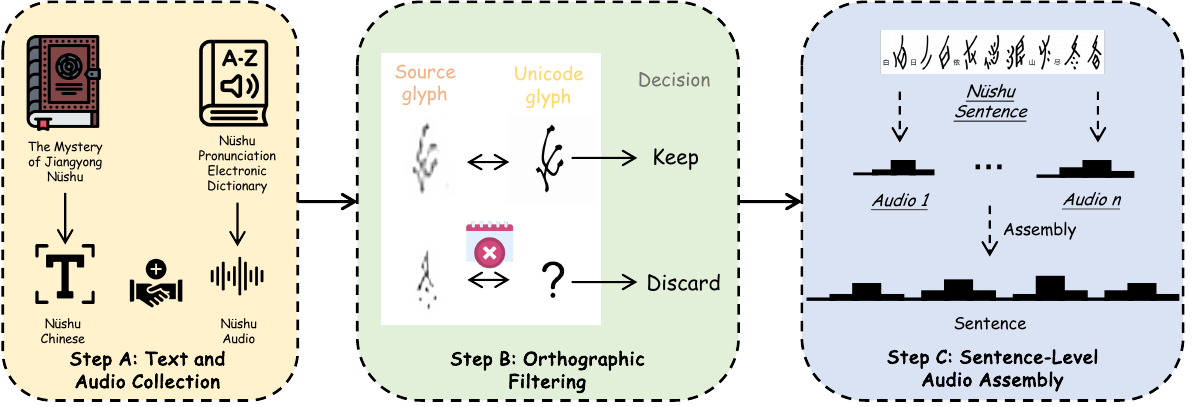}
    \caption{Overview of the construction pipeline of \textbf{NüshuVoice}. The pipeline consists of three stages: data collection, rigorous orthographic filtering, and sentence-level audio assembly.}
    \label{fig:nushuspeech_construction}
\end{figure*}

Our contributions are summarized as follows:
\begin{itemize}[itemsep=0pt, topsep=2pt, parsep=0pt, partopsep=0pt]
    \item We construct \textbf{NüshuVoice}, a sentence-level multimodal Nüshu TTS dataset by aligning standardized Nüshu Unicode text, phonetic transcriptions, Chinese translations, and archival syllable-level recordings.
    \item We propose \textbf{Nüshu-PitchVITS}, an pitch-conditioned VITS framework that leverages the five-level pitch notation of Nüshu as an explicit prosodic inductive bias.
    \item We provide the first systematic objective and subjective evaluation of neural TTS models for Nüshu, demonstrating the effectiveness of explicit pitch modeling in an extreme low-resource setting.
    \item We contribute to the computational preservation and acoustic revitalization of Nüshu by making its pronunciation accessible for future research, documentation, and cultural heritage applications.
\end{itemize}

\section{Related Work}
\subsection{Digital Preservation and Computational Studies of Nüshu}
While Nüshu has recently garnered increasing cultural and linguistic interest \cite{thurnell2022introduction}, computational research on this endangered script remains limited. Foundational preservation efforts have mainly focused on linguistic documentation and digitization \cite{rao2022intangible}. Notably, Xie Zhimin compiled, translated, and provided phonetic annotations for surviving Nüshu manuscripts \footnote{Xie, Zhimin. \textit{Jiangyong Nüshu zhi mi} [\textit{The Mystery of Jiangyong Nüshu}]. Zhengzhou: Henan Renmin Chubanshe [Henan People's Publishing House], 1991, pp. 700--709. In Chinese.}, while the electronic dictionary developed by Xie Zhimin and Wang Lihua digitized 873 isolated syllable recordings, preserving the authentic pronunciations of fluent speakers \footnote{Xie, Zhimin, and Wang, Lihua. \textit{Nüshu Fasheng Dianzi Zidian} [\textit{Nüshu Pronunciation Electronic Dictionary}]. Wuhan: Huazhong University of Science and Technology Press, 2002. In Chinese.}.

Building upon these linguistic resources, the machine learning and NLP communities have recently initiated exploratory studies on Nüshu. Early interdisciplinary work by \citet{sun2023ai} introduced an AI-driven approach to conceptualize a standardized Nüshu encoding system. More recently, computational efforts have shifted toward specific modalities. In the text domain, \citet{yang2024nushurescue} proposed a framework for leveraging large language models on extremely low-resource endangered languages like Nüshu. In the visual domain, \citet{yang2025recontextualizing} established foundational optical character recognition (OCR) benchmarks by constructing specialized visual datasets and fine-tuning vision-based recognition models. Despite these emerging advancements in text generation and visual recognition, the acoustic modality of Nüshu remains largely unexplored. 

\subsection{Text-to-Speech for Low-Resource Languages}




Neural text to speech systems usually require substantial paired text and speech data, which are unavailable for many low resource languages and endangered writing traditions \cite{xu2020lrspeech, park2023unsupervised}. However, recent studies often define low resource conditions in a less restrictive way than the setting considered in this work. For example, \citet{kwon2025parameter} studied cross lingual continuous fine tuning with a 12 hour Korean corpus. \citet{basher2025bntts} achieved few shot adaptation with 4.22 hours of high quality target speech, supported by 3.85k hours of continuous pretraining. These settings still assume several hours of clean target data or large scale external acoustic priors. They are therefore not directly applicable to Nüshu, where the available recordings are extremely limited, archival, and non studio in quality.

This constraint makes the choice of TTS architecture critical. Autoregressive models such as Tacotron 2 \cite{shen2018natural} can produce natural speech, but their attention mechanisms are prone to collapse when trained with limited data. Non autoregressive models such as FastSpeech 2 \cite{ren2020fastspeech} are more stable, but they usually require duration labels or external alignment supervision. Recent flow matching and zero shot models such as F5 TTS \cite{chen2025f5} show strong generalization, but their performance depends on large scale acoustic pretraining. In contrast, VITS \cite{kim2021conditional} is a more suitable base model for this setting because it is trained end to end and uses Monotonic Alignment Search to learn text and speech alignment without an external aligner.

Standard VITS is still not sufficient because its unsupervised alignment process can become unstable when the paired data are too limited. We therefore introduce explicit fundamental frequency conditioning. $F0$ provides direct pitch information and acts as a prosodic inductive bias which helps regularize the acoustic latent representation. This design is also consistent with the structure of the Nüshu data, since the corpus uses a five level pitch notation \cite{zhengzhang2015wu}. $F0$ conditioning is therefore both an architectural choice for stabilizing VITS and a data driven choice that reflects the available Nüshu annotation.

\begin{table*}[t]
  \centering
  \includegraphics[width=\textwidth]{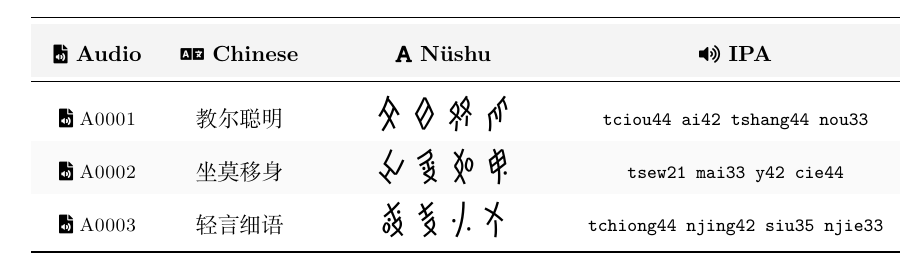}
  \caption{Some examples of \textbf{NüshuVoice} showing the multimodal alignment.}
  \label{tab:dataset_examples}
\end{table*}

\section{NüshuVoice Construction}
As illustrated in \textbf{Figure~\ref{fig:nushuspeech_construction}}, this section details the construction of our Nüshu text-to-audio dataset, including data collection, rigorous orthographic filtering and audio assembly.

\subsection{Text and Audio Collection}
The foundation of our dataset relies on two primary historical and linguistic resources. The textual corpus is derived from \textit{The Mystery of Jiangyong Nüshu}.\footnote{Xie, Zhimin. \textit{Jiangyong Nüshu zhi mi} [\textit{The Mystery of Jiangyong Nüshu}]. Zhengzhou: Henan Renmin Chubanshe [Henan People's Publishing House], 1991, pp. 700--709. In Chinese.} This comprehensive compilation contains approximately 63,665 characters across various literary forms, including original letters, narrative poems, and folk songs. Crucially, this resource provides the original Nüshu characters alongside their phonetic transcriptions and standard Chinese translations at the sentence level. For the acoustic modality, we extracted raw audio data from the \textit{Nüshu Pronunciation Electronic Dictionary}.\footnote{Xie, Zhimin, and Wang, Lihua. \textit{Nüshu Fasheng Dianzi Zidian} [\textit{Nüshu Pronunciation Electronic Dictionary}]. Wuhan: Huazhong University of Science and Technology Press, 2002. In Chinese.} This dictionary contains 873 isolated syllable recordings in WAV format, capturing the pronunciation of individual Nüshu characters directly from fluent speakers.

\subsection{Rigorous Orthographic Filtering}

Given the highly idiosyncratic and historically non-standardized nature of Nüshu writing, establishing a reliable mapping between archival Nüshu texts and modern Unicode representations is essential. We therefore implemented a strict visual and semantic cross-referencing protocol to map handwritten and printed Nüshu glyphs from the 1991 manuscript to their corresponding standardized Unicode forms.

Annotators manually verified every character within each candidate sentence. To ensure orthographic fidelity, we adopted a conservative filtering strategy: if any character in a sentence lacked a corresponding Unicode equivalent, or exhibited even minor structural discrepancies, such as a missing stroke or a variant component, the entire sentence was discarded. Sentences that passed this filtering stage were stored in a structured metadata table containing the standardized Nüshu Unicode string, phonetic transcription, standard Chinese translation, and sentence identifier.

\subsection{Sentence-Level Audio Assembly}
Since the primary acoustic resource only offers character-level pronunciations, we developed a deterministic synthesis pipeline to generate continuous, sentence-level audio. Relying on the precise phonetic transcriptions in the original corpus, which utilize a five-level tone marking system\footnote{Unlike standard Mandarin Pinyin, Nüshu phonetic notation employs the International Phonetic Alphabet (IPA) alongside a five-level pitch scale to capture the local dialect.}, we mapped each verified Unicode character to its corresponding isolated WAV file from the electronic dictionary. These character-level audio segments were then temporally concatenated via Audacity to reconstruct the complete utterance. Finally, the synthesized audio files were assigned unique sentence identifiers, establishing a one-to-one multimodal alignment between text and audio. Representative examples of the resulting multimodal alignment are shown in \textbf{Table~\ref{tab:dataset_examples}}.  The statistical summary is showed in \textbf{Table~\ref{tab:statisticalsummaryofdataset}}.

\begin{table}[t]
\small
\centering
\label{tab:dataset_stats}
\renewcommand{\arraystretch}{1.2} 
\setlength{\tabcolsep}{8pt}      
\begin{tabular}{lcc}              
\toprule
\rowcolor{TableHeaderGray} 
\textbf{Data Split} & \textbf{Utterances} & \textbf{Duration (mins)} \\ 
\midrule
Train               & 1,002               & 79.00                    \\
Validation          & 125                 & 10.00                    \\
Test                & 125                 & 9.80                     \\ 
\midrule
\rowcolor{OurMethodBlue}
\textbf{Total}      & \textbf{1,252}      & \textbf{98.80}           \\ 
\bottomrule
\end{tabular}
\caption{Statistical summary of the \textbf{NüshuVoice}.}
\label{tab:statisticalsummaryofdataset}
\end{table}

\subsection{Expert Validation}

To assess the reliability of NüshuVoice, we randomly sampled 10\% of the dataset, i.e., 125 out of 1,252 entries, and invited two Nüshu domain experts to independently validate them. The experts examined glyph--Unicode consistency, phonetic transcriptions, Chinese translations, and audio--syllable alignment. The two experts achieved 90.00\% agreement, with a Cohen's $\kappa$ of 0.7887, indicating substantial agreement. After adjudication, 117 entries were retained as correct, corresponding to a correctness rate of 93.6\%, suggesting the high overall quality of NüshuVoice.

\section{Proposed Nüshu-PitchVITS}

\begin{figure*}[h!]
    \centering
    \includegraphics[width=\textwidth]{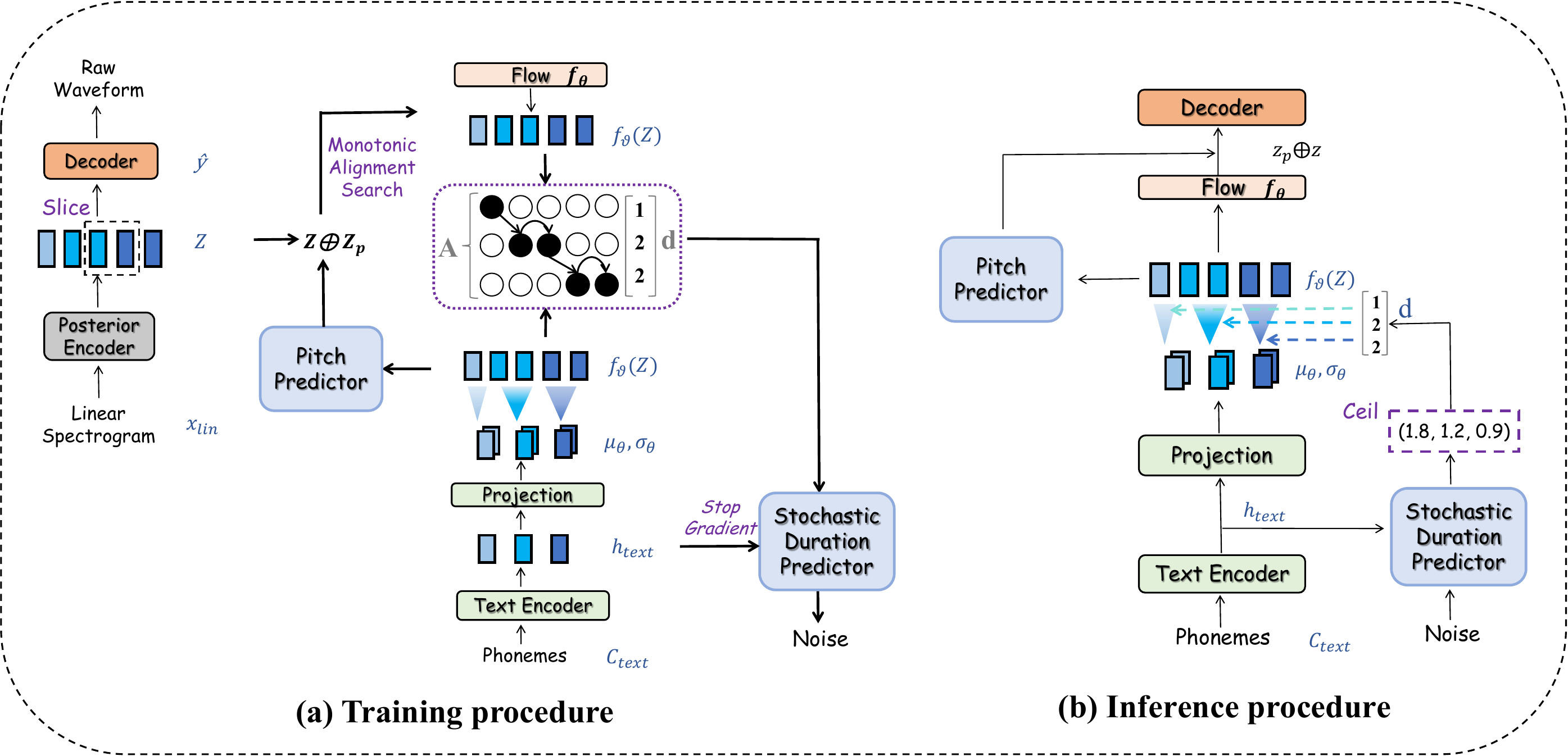}
    \caption{
    Overview of the proposed \textbf{Nüshu-PitchVITS} framework.
    (a) During training, the text encoder produces phoneme-level hidden states,
    which are expanded to the frame level by MAS. A pitch predictor estimates
    the frame-level log-$F_0$ sequence and injects the projected pitch
    representation into the latent acoustic representation before waveform
    decoding. The model is optimized with reconstruction, KL-divergence,
    duration, pitch prediction, adversarial, and feature matching losses.
    (b) During inference, the predicted log-$F_0$ is used together with the
    text-derived latent representation to generate the final waveform.
    }
    \label{fig:train_infer}
\end{figure*}

\subsection{Overview}

As shown in \textbf{Figure~\ref{fig:train_infer}}, Nüshu-PitchVITS extends the original VITS \cite{kim2021conditional} architecture by incorporating an explicit fundamental frequency ($F_0$) predictor. This design is motivated by the phonological characteristics of N\"ushu, which employs a five-level pitch transcription system representing tone values from 1 to 5.
\subsection{Revisit: VITS}

VITS \cite{kim2021conditional} can be formulated as a conditional
variational autoencoder (VAE). Given a target waveform $\mathbf y$ and a
condition $\mathbf c$, its objective is to maximize
$\log p_\theta(\mathbf y \mid \mathbf c)$ via the ELBO:
\begin{equation}
\begin{aligned}
\log p_\theta(\mathbf y \mid \mathbf c)
&\geq
\E_{q_\phi}
\left[
    \log p_\theta(\mathbf y \mid \mathbf Z)
\right]  \\
&\quad -
\KL
\left(
    q_\phi(\mathbf Z \mid \mathbf X_{\mathrm{lin}})
    \,\|\, 
    p_\theta(\mathbf Z \mid \mathbf c)
\right).
\end{aligned}
\label{eq:vits_elbo}
\end{equation}
where $\mathbf Z$ is the latent acoustic representation and
$\mathbf X_{\mathrm{lin}}$ is the linear-scale spectrogram used by the
posterior encoder.

\subsection{Nüshu-PitchVITS Architecture}

The input to Nüshu-PitchVITS is the phonetic transcription of a Nüshu sentence, where each syllabic unit includes its five-level tone value. This tone-aware representation is encoded by the text encoder and serves as the basis for both acoustic alignment and pitch prediction.

To model the tonal characteristics of Nüshu, we augment VITS with a
frame-level pitch prediction branch. Let the text encoder output be
\begin{equation}
    \mathbf H =
    [\mathbf h_1,\ldots,\mathbf h_N]^\top
    \in \R^{N\times d},
    \label{eq:text_hidden}
\end{equation}
where $N$ is the phoneme length and $d$ is the hidden dimension.

After Monotonic Alignment Search (MAS), we obtain a hard monotonic
alignment matrix
\begin{equation}
    \mathbf A^* \in \{0,1\}^{T\times N},
    \qquad
    \sum_{n=1}^{N} A^*_{t,n}=1,
    \label{eq:alignment}
\end{equation}
where $T$ is the number of acoustic frames. The phoneme-level hidden states
are expanded to the frame level by
\begin{equation}
    \tilde{\mathbf H}
    =
    \mathbf A^* \mathbf H
    \in \R^{T\times d}.
    \label{eq:expanded_hidden}
\end{equation}

The pitch predictor estimates the frame-level log-$F_0$ sequence:
\begin{equation}
    \hat{\mathbf f}
    =
    g_\psi(\tilde{\mathbf H}),
    \qquad
    \hat{\mathbf f}\in\R^T,
    \label{eq:pitch_predictor}
\end{equation}
where $g_\psi(\cdot)$ is a lightweight two-layer 1D convolutional network
with layer normalization and ReLU activation.

The predicted log-$F_0$ sequence is projected to the hidden dimension and
added to the latent representation:
\begin{equation}
    \mathbf E_{F_0}
    =
    W_{F_0}(\hat{\mathbf f})
    \in \R^{T\times d},
    \qquad
    \mathbf Z'=\mathbf Z+\mathbf E_{F_0}.
    \label{eq:pitch_injection}
\end{equation}
Here, $W_{F_0}$ denotes a frame-wise linear projection from
$\R$ to $\R^d$. The waveform decoder then generates
\begin{equation}
    \hat{\mathbf y}=G_\theta(\mathbf Z').
    \label{eq:decoder}
\end{equation}

To avoid disturbing alignment estimation in the low-resource setting, the
pitch branch is excluded from MAS. The alignment $\mathbf A^*$ is computed
only from the prior statistics and the flow-transformed posterior latent
variables. Thus, the pitch predictor affects acoustic realization only after
the alignment has been determined.

For supervision, the ground-truth fundamental frequency sequence
$\mathbf F_0\in\R^T$ is extracted from the target waveform using the DIO
algorithm in WORLD and converted to the logarithmic domain:
\begin{equation}
    \mathbf f
    =
    \log(\mathbf F_0+\epsilon),
    \qquad
    \mathbf f\in\R^T,
    \label{eq:log_f0}
\end{equation}
where $\epsilon$ is a small positive constant. Since both $\mathbf f$ and
$\hat{\mathbf f}$ are log-$F_0$ sequences, the pitch loss is
\begin{equation}
    \LFzero
    =
    \frac{1}{T}
    \left\|
        \mathbf f-\hat{\mathbf f}
    \right\|_2^2.
    \label{eq:f0_loss}
\end{equation}

\begin{algorithm}[t]
\footnotesize

\KwInput{
Tone-aware phonetic sequence $\mathbf c$; target waveform $\mathbf y$; 
training stage $s \in \{1,2\}$
}
\KwOutput{
Synthesized waveform $\hat{\mathbf y}$
}
\KwInit{
Pre-trained VITS parameters $\Theta$; pitch predictor $g_\psi$; 
pitch projection $W_{F_0}$
}

\BlankLine
\textbf{Training phase}\;

Encode $\mathbf c$ into phoneme-level hidden states:
$\mathbf H \leftarrow \mathrm{TextEnc}_{\Theta}(\mathbf c)$\;

Extract posterior latent representation:
$\mathbf Z \leftarrow \mathrm{PostEnc}_{\Theta}(\mathbf y)$\;

Estimate monotonic alignment path using MAS:
$\mathbf A^* \leftarrow \mathrm{MAS}(\mathbf Z, \mathbf H)$\;

Expand hidden states to the acoustic frame level:
$\widetilde{\mathbf H} \leftarrow \mathbf A^* \mathbf H$\;

Predict frame-level log-pitch:
$\hat{\mathbf f} \leftarrow g_\psi(\widetilde{\mathbf H})$\;

Inject pitch information into the latent representation:
$\mathbf Z' \leftarrow \mathbf Z + W_{F_0}(\hat{\mathbf f})$\;

Generate waveform:
$\hat{\mathbf y} \leftarrow G_{\Theta}(\mathbf Z')$\;

Extract ground-truth log-$F_0$ from $\mathbf y$ using WORLD-DIO:
$\mathbf f \leftarrow \log(F_0(\mathbf y)+\epsilon)$\;

\BlankLine
\If{$s=1$}{
Freeze posterior encoder and waveform decoder\;
Update text encoder, duration predictor, flow, and pitch predictor using
$\Lkl + \Ldur + \LFzero$\;
}
\ElseIf{$s=2$}{
Unfreeze all modules\;
Update the full model using
$\mathcal L_G =
\Ladv + \lambda_{\mathrm{fm}}\Lfm
+ \lambda_{\mathrm{mel}}\Lrecon
+ \lambda_{\mathrm{dur}}\Ldur
+ \lambda_{\mathrm{kl}}\Lkl
+ \lambda_{F_0}\LFzero$\;
}

\BlankLine
\textbf{Inference phase}\;

Predict duration and alignment from $\mathbf c$\;
Predict $\hat{\mathbf f}$ from the expanded text representation\;
Construct pitch-conditioned latent representation $\mathbf Z'$\;
Decode $\mathbf Z'$ to synthesize $\hat{\mathbf y}$\;

\caption{\textbf{Nüshu-PitchVITS}: pitch-aware training and inference}
\label{alg:nushu_pitchvits}
\end{algorithm}

\subsection{Two-stage Training}

To accelerate convergence and ensure stable optimization, we employ a
two-stage training strategy initialized from a pre-trained English VITS
model. In the first stage, we freeze the waveform decoder and posterior
encoder, and train only the text encoder, pitch predictor, duration
predictor, and normalizing flow. This stage is optimized using $\Lkl$,
$\Ldur$, and $\LFzero$. In the second stage, we unfreeze all modules and
perform end-to-end joint training.

\subsection{Training Objective}

The total generator loss is
\begin{equation}
\begin{aligned}
\mathcal L_G
=
&\Ladv(G)
+\lambda_{\mathrm{fm}}\Lfm(G)
+\lambda_{\mathrm{mel}}\Lrecon  \\
&+\lambda_{\mathrm{dur}}\Ldur
+\lambda_{\mathrm{kl}}\Lkl
+\lambda_{F_0}\LFzero ,
\end{aligned}
\label{eq:total_loss}
\end{equation}
where the $\lambda$ terms are loss weights.

The mel-spectrogram reconstruction loss is
\begin{equation}
    \Lrecon
    =
    \left\|
        \mathbf X_{\mathrm{mel}}
        -
        \hat{\mathbf X}_{\mathrm{mel}}
    \right\|_1 .
    \label{eq:mel_loss}
\end{equation}

The KL-divergence loss is defined as
\begin{equation}
    \Lkl
    =
    \KL
    \big(
        q_\phi(\mathbf Z \mid \mathbf X_{\mathrm{lin}})
        \|
        p_\theta(\mathbf Z \mid \mathbf c_{\mathrm{text}},\mathbf A^*)
    \big),
    \label{eq:kl_loss}
\end{equation}
where $\mathbf c_{\mathrm{text}}$ is the text condition and
$q_\phi(\mathbf Z \mid \mathbf X_{\mathrm{lin}})$ is a diagonal Gaussian
posterior.

The adversarial losses follow the least-squares GAN objective in VITS:
\begin{align}
    \Ladv(D)
    &=
    \sum_{k=1}^{K}
    \E
    \left[
        (D_k(\mathbf y)-1)^2
        +
        D_k(\hat{\mathbf y})^2
    \right],
    \label{eq:adv_d}
    \\
    \Ladv(G)
    &=
    \sum_{k=1}^{K}
    \E
    \left[
        (D_k(\hat{\mathbf y})-1)^2
    \right],
    \label{eq:adv_g}
\end{align}
where $\{D_k\}_{k=1}^{K}$ denotes the set of discriminators.

The feature matching loss is
\begin{equation}
    \Lfm(G)
    =
    \sum_{k,\ell}
    \frac{1}{N_{k,\ell}}
    \E
    \left[
        \left\|
        D_{k,\ell}(\mathbf y)
        -
        D_{k,\ell}(\hat{\mathbf y})
        \right\|_1
    \right],
    \label{eq:fm_loss}
\end{equation}
where $D_{k,\ell}(\cdot)$ is the $\ell$-th feature map of the $k$-th
discriminator, and $N_{k,\ell}$ is the number of elements in that feature
map.


\begin{table*}[t]
\centering
\small
\setlength{\tabcolsep}{5pt}
\renewcommand{\arraystretch}{1.15}
\begin{tabular}{@{}lccccc@{}}
\toprule
\multirow{2}{*}[-0.5ex]{\textbf{Model}} 
& \multicolumn{3}{c}{\textbf{Objective Metrics}} 
& \multicolumn{2}{c}{\textbf{Subjective MOS}} \\
\cmidrule(lr){2-4} \cmidrule(lr){5-6}
& \textbf{MCD} $\downarrow$ 
& \textbf{$F_0$ RMSE} $\downarrow$ 
& \textbf{$F_0$ Corr.} $\uparrow$ 
& \textbf{Naturalness} $\uparrow$ 
& \textbf{Intelligibility} $\uparrow$ \\
\midrule
Tacotron 2   & \res{44.26}{4.34} & \res{47.15}{16.01} & \res{0.32}{0.38} & \res{1.66}{0.84} & \res{1.03}{0.18} \\
FastSpeech 2 & \res{37.53}{5.91} & \res{29.51}{13.06} & \res{0.66}{0.27} & \res{1.90}{0.71} & \res{1.10}{0.30} \\
F5-TTS       & \res{9.96}{1.92}  & \res{37.03}{18.70} & \res{0.47}{0.26} & \res{3.10}{0.92} & \res{3.27}{0.44} \\
Glow-TTS     & \res{39.96}{3.91} & \res{28.44}{13.77} & \res{0.72}{0.22} & \res{2.61}{0.82} & \res{1.03}{0.18} \\
VITS         & \res{5.19}{1.28}  & \res{23.42}{13.13} & \res{0.73}{0.31} & \res{3.03}{0.56} & \res{3.90}{0.55} \\
\midrule
\textbf{Nüshu-PitchVITS} 
& \best{3.11}{1.34}
& \best{13.23}{9.63}
& \best{0.92}{0.12}
& \best{3.31}{0.53}
& \best{4.97}{0.18} \\
\midrule
Ground Truth 
& -- & -- & -- 
& \res{3.34}{0.60}
& N/A$^{\dagger}$ \\
\bottomrule
\end{tabular}
\caption{Objective and subjective evaluation results on \textbf{NüshuVoice}. Objective metrics include Mel-Cepstral Distortion (MCD), $F_0$ RMSE, and $F_0$ correlation. Subjective metrics include naturalness and intelligibility MOS. The best model results are shown in \textbf{bold}. $^{\dagger}$Ground-truth intelligibility MOS was not reported.}
\label{tab:main_results}
\end{table*}

\section{Experiments}
\subsection{Implementation Details}

We initialized Nüshu-PitchVITS using the weights of a pre-trained English VITS model \cite{kim2021conditional}, rather than training all parameters from scratch. The model was then trained with our two-stage optimization strategy on \textbf{NüshuVoice}. All training procedures were conducted on a single NVIDIA RTX 5090 GPU. To improve computational efficiency, we used mixed-precision training (FP16) and a data loading pipeline with four worker processes. We used AdamW optimizer with an initial learning rate of $1\times10^{-4}$ in the first training stage and reduced it to $5\times10^{-5}$ during second.

\subsection{Baselines}

We compare Nüshu-PitchVITS with five representative TTS baselines: Tacotron 2 \cite{shen2018natural}, FastSpeech 2 \cite{ren2020fastspeech}, Glow-TTS \cite{kim2020glow}, F5-TTS \cite{chen2025f5}, and standard VITS \cite{kim2021conditional}. These models cover major neural TTS paradigms, including autoregressive sequence-to-sequence synthesis, non-autoregressive duration-based synthesis, flow-based synthesis, flow-matching-based synthesis, and end-to-end variational adversarial synthesis. For a fair comparison, all baseline models were trained and evaluated on the same NüshuVoice train, validation, and test splits.

\subsection{Evaluation Metrics}

To comprehensively evaluate the quality of the synthesized Nüshu speech, we employ a combination of objective and subjective metrics. Objective assessments include Mel-Cepstral Distortion (MCD) \cite{kubichek1993mel} to measure spectral fidelity. Furthermore, given the critical role of pitch guidance in our proposed architecture, we compute the Root Mean Square Error (RMSE) and the Pearson correlation coefficient of the fundamental frequency ($F_0$) to evaluate prosodic reconstruction accuracy. For subjective evaluation, we conduct Mean Opinion Score (MOS) tests to assess the generated audio across two dimensions: naturalness and intelligibility \cite{international1996methods}. 

Notably, while Word Error Rate (WER) computed via an off-the-shelf Automatic Speech Recognition (ASR) model is the standard objective proxy for intelligibility in conventional TTS studies, the complete absence of Nüshu ASR systems precludes this approach. Consequently, we rely on human-evaluated intelligibility MOS as a necessary alternative to assess the linguistic clarity and pronunciation correctness of the synthesized outputs. Detailed metric definitions and the subjective evaluation guidelines are provided in Appendix~\ref{sec:appendix_eval}.


\section{Results}

\subsection{Objective TTS Evaluation}
The objective evaluation results, detailed in \textbf{Table \ref{tab:main_results}}, clearly demonstrate the fragility of conventional TTS architectures under extreme low-resource constraints. Traditional models, including Tacotron 2, FastSpeech 2, and Glow-TTS, suffer catastrophic degradation, exhibiting severe spectral distortion (MCD > 37) and virtually no pitch correlation. While the standard VITS model emerges as the strongest baseline, our proposed method achieves substantial improvements across all metrics. By explicitly integrating an $F_0$ predictor, our framework minimizes the MCD to 3.11 and drastically reduces the $F_0$ RMSE to 13.23. Furthermore, it achieves a highly accurate $F_0$ correlation of 0.92, confirming that explicit prosodic guidance effectively stabilizes acoustic generation and prevents spectral collapse.


\subsection{Subjective Human Evaluation}
The subjective Mean Opinion Scores (\textbf{Table \ref{tab:main_results}}) further corroborate our objective findings. Standard baselines fail entirely to produce comprehensible Nüshu speech, yielding Intelligibility MOS scores near 1.0 (indicating unintelligible babbling). Although recent models like F5 and VITS manage to generate partially recognizable audio, our method significantly surpasses them, achieving a near-perfect Intelligibility MOS of 4.97. Crucially, our model attains a Naturalness MOS of 3.31, performing highly competitively with the ground-truth recordings (3.34). These results conclusively demonstrate that our approach not only preserves the strict phonetic integrity of the Nüshu script but also reconstructs its authentic tonal dynamics.

\subsection{Spectral Analysis}
\begin{figure}[htbp]
    \large
    \centering
    \includegraphics[width=\linewidth]{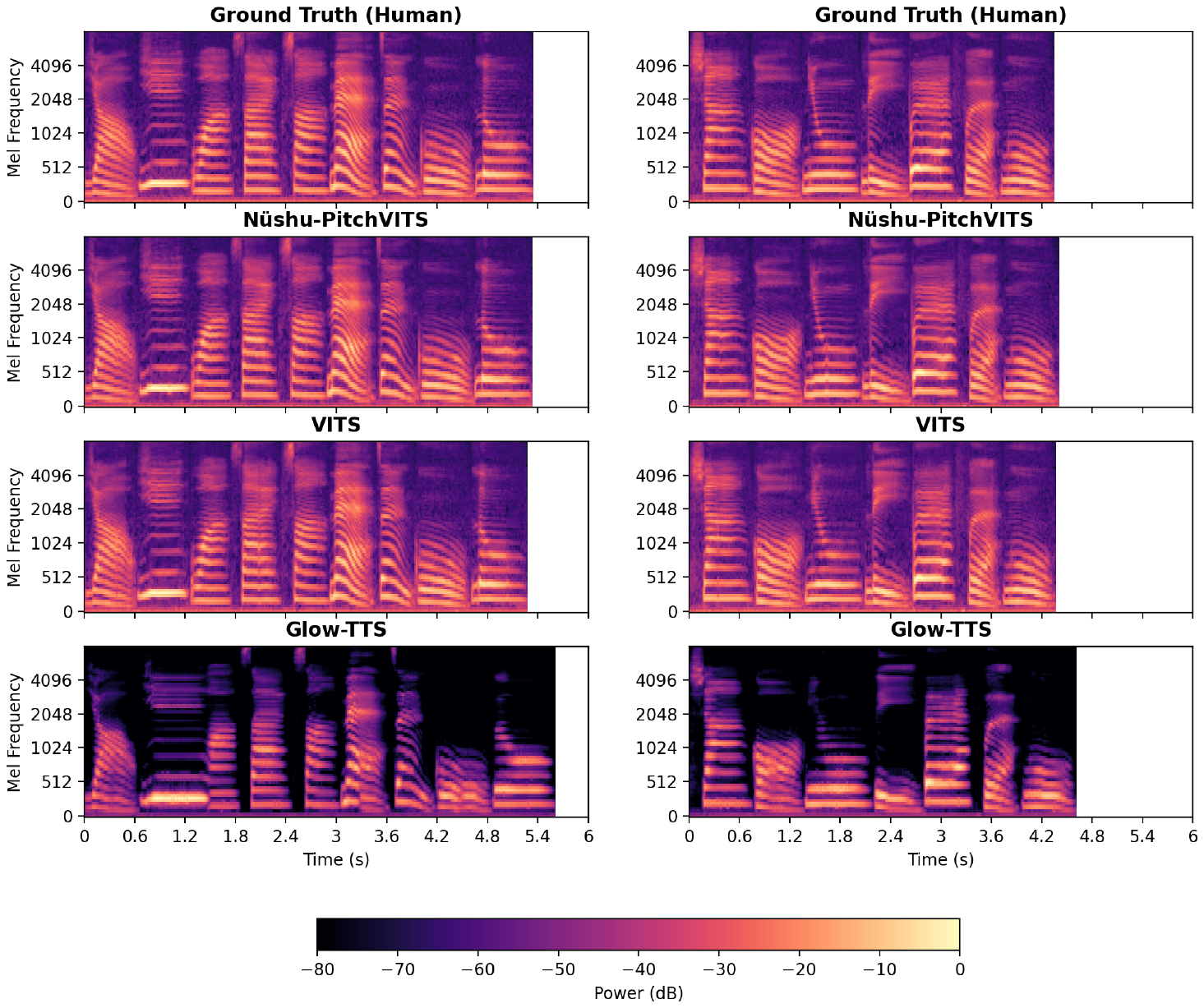}
    \caption{Mel-spectrogram comparison of two Nüshu utterances. 
    The left column corresponds to an utterance with IPA transcription 
    [iou13 i5 va33 si44 suow44 mai42 tseng42 ku21 lu21], and the right 
    column corresponds to [ciang13 kou21 njyu33 tie42 pwe5 fwe13 ngu13].}
    \label{fig:mel_comparison}
\end{figure}

As shown in \textbf{Figure~\ref{fig:mel_comparison}}, both VITS and our Nüshu-PitchVITS generate recognizable harmonic structures and clear formant trajectories, suggesting that VITS-style end-to-end alignment is relatively robust in the Nüshu low-resource setting. However, standard VITS still exhibits noticeable over-smoothing, weakened harmonic contrast, and less stable energy distribution, especially around syllable transitions. In contrast, Nüshu-PitchVITS produces spectrograms that are closer to the ground truth, with sharper harmonic bands, smoother temporal continuity, and more consistent pitch-related structures. These differences indicate that the explicit $F_0$ branch helps the model preserve tonal dynamics and improves prosodic stability beyond what can be achieved by standard latent acoustic modeling alone. 

\section{Conclusion}

This paper presents the first dedicated study on text-to-audio modeling for Nüshu, an endangered phonetic script with unique linguistic and cultural significance. To address the lack of acoustic resources, we constructed a sentence-level Nüshu TTS dataset by aligning verified Unicode Nüshu text, phonetic transcriptions, standard Chinese translations, and archival recordings. 

Building on this dataset, we proposed Nüshu-PitchVITS tailored to the extreme low-resource characteristics of Nüshu. By introducing an explicit pitch prediction branch, the model incorporates tonal information as a prosodic inductive bias. Experimental results show that the proposed method consistently outperforms strong TTS baselines in spectral fidelity, pitch reconstruction, and human perceptual evaluation.

Beyond model performance, this work highlights the importance of acoustic preservation for endangered writing traditions. Since Nüshu is inherently phonetic and closely tied to the Jiangyong dialect, recovering its pronunciation is essential for preserving not only its written form but also its oral and cultural heritage. We hope that the proposed dataset and modeling framework can serve as a foundation for future research on low-resource TTS, endangered language revitalization, and computational approaches to intangible cultural heritage.

\section*{Limitations}

This work has several limitations. First, our sentence-level audio is assembled from isolated syllable recordings rather than recorded as natural continuous speech. As a result, the dataset cannot fully capture coarticulation, pauses, rhythm, and discourse-level prosody that would appear in spontaneous or naturally read utterances. Second, the scale of the dataset is inherently constrained by the scarcity of surviving Nüshu acoustic resources. Although NüshuVoice contains 1,252 utterances and 98.8 minutes of audio, this amount necessarily limits the diversity of speakers, speaking styles, and phonetic contexts available for model training and evaluation. Third, because sentence-level audio is assembled from a finite inventory of archival syllable recordings, train and test splits may share underlying acoustic units. Therefore, objective results should be interpreted within this controlled archival reconstruction setting. Finally, our strict Unicode-based filtering improves annotation reliability, but it also excludes historical glyph variants without exact digital equivalents.

\section*{Ethics Considerations}

Nüshu is an endangered script and an important form of women's cultural heritage. This work aims to support preservation and linguistic documentation, not to replace community knowledge or living cultural practice. Since the audio comes from archival pronunciation resources, any release of derived data, models, or generated samples should respect the licensing conditions and rights associated with the original recordings. Synthetic speech should be clearly labeled as machine-generated to avoid cultural misrepresentation or confusion with authentic historical recordings. Future work should involve Nüshu scholars, local cultural practitioners, and community stakeholders.

\section*{Declaration of AI Usage}

The authors used AI-assisted writing tools to improve the clarity, grammar, and readability of the manuscript. All AI-generated suggestions were carefully reviewed, edited, and verified by the authors. The authors take full responsibility for the content of the paper, including the proposed method, dataset construction, experimental design, results, and conclusions.


\bibliography{custom}

\appendix

\section{Additional Dataset Construction Details}
\label{sec:appendix_dataset}

This appendix provides additional details on the construction of
NüshuVoice, including archival resource recovery, Unicode-based
orthographic filtering, sentence-level audio assembly, and dataset
organization. These details complement the main dataset construction
section and are intended to improve reproducibility.

\subsection{Archival Resource Recovery}

The construction of NüshuVoice relies on two complementary archival
resources. The textual source is \textit{The Mystery of Jiangyong Nüshu},
which provides Nüshu sentences together with standard Chinese translations
and phonetic transcriptions. The acoustic source is the \textit{Nüshu
Pronunciation Electronic Dictionary}, which contains isolated
syllable-level pronunciation recordings of Nüshu characters.

Because the electronic dictionary was originally released as a CD-ROM
resource, its software interface is not reliably executable on modern
operating systems. We therefore recovered the raw acoustic data directly
from the original CD-ROM contents. Specifically, we extracted the isolated
WAV recordings from the \texttt{voice} directory and used only these
original pronunciation files for sentence-level audio construction. This
step was necessary because no modern large-scale Nüshu speech corpus is
available, and the recovered recordings constitute a rare acoustic resource
for preserving Nüshu.

\subsection{Digital Environment and Lookup Tools}

Dataset construction was conducted in a Windows environment with Nüshu font
and input support. We used the Rime Nüshu input method to input and render
standardized Unicode Nüshu characters. For candidate character lookup and
Unicode conversion, we referred to the online converter associated with the
\textit{Nüshu Standard Character Calligraphy Copybook}. Audio editing,
concatenation, and waveform export were performed using Audacity.

These tools were used only for rendering, lookup, and audio editing. All
final decisions about character validity were made through manual
verification against the archival text rather than automatic conversion
alone.

\subsection{Unicode-Based Orthographic Filtering}

Historical Nüshu glyphs are not fully standardized, and the same syllable
may appear in different handwritten or printed variants. To build a reliable
Unicode-based TTS dataset, we adopted a conservative sentence-level
filtering strategy. Each candidate sentence was manually inspected
character by character. A sentence was retained only when every source glyph
had a visually and structurally consistent Unicode counterpart.

If any character in a sentence lacked a Unicode equivalent, showed an
ambiguous glyph shape, or differed from the corresponding Unicode form in
stroke structure or component composition, the entire sentence was
discarded. This strict filtering strategy reduces coverage but improves the
orthographic reliability of the final dataset.

\begin{table*}[t]
\centering
\small
\setlength{\tabcolsep}{5pt}
\renewcommand{\arraystretch}{1.15}
\begin{tabular}{p{0.42\linewidth}p{0.45\linewidth}}
\toprule
\textbf{Case} & \textbf{Decision} \\
\midrule
The source glyph has an exact Unicode counterpart. 
& Retain the character. \\

All characters in the sentence have visually consistent Unicode forms. 
& Retain the sentence. \\

A source glyph has no corresponding Unicode character. 
& Discard the entire sentence. \\

A source glyph differs from the Unicode form by a missing or additional stroke. 
& Discard the entire sentence. \\

A source glyph contains a variant component or structurally different radical. 
& Discard the entire sentence. \\

The glyph--Unicode mapping is visually ambiguous. 
& Discard the entire sentence. \\
\bottomrule
\end{tabular}
\caption{Unicode-based orthographic filtering criteria used in the construction of NüshuVoice.}
\label{tab:unicode_filtering}
\end{table*}

\subsection{Sentence-Level Audio Assembly}

The acoustic dictionary provides isolated syllable-level recordings rather
than continuous sentence-level speech. Therefore, we constructed
sentence-level audio using a deterministic concatenation procedure. For each
retained sentence, we first used the phonetic transcription in the textual
source to identify the pronunciation of each Nüshu character. We then
matched each phonetic unit to its corresponding isolated WAV file in the
recovered electronic dictionary.

The selected syllable recordings were arranged in sentence order and
concatenated using Audacity to form a complete utterance. Each generated
waveform was exported as a sentence-level audio file and assigned the same
unique identifier as its corresponding metadata entry. This procedure
creates a one-to-one alignment between standardized Nüshu text, phonetic
transcription, Chinese translation, and sentence-level audio.

\subsection{Dataset Organization}

The final dataset consists of a metadata table and a set of waveform files.
Each metadata entry contains a unique sentence identifier, standardized
Unicode Nüshu text, phonetic transcription, and standard Chinese
translation. The waveform folder contains the corresponding sentence-level
audio files. The shared sentence identifier links each textual entry to its
audio file.

This organization enables both text-to-speech training and multimodal
analysis. It also makes the construction process transparent: each
sentence-level audio file can be traced back to its Unicode Nüshu text,
phonetic transcription, and the sequence of isolated archival syllable
recordings used to assemble it.

\section{Evaluation Details}
\label{sec:appendix_eval}

This appendix provides additional details on the objective metric computation and subjective evaluation protocol used in our experiments.

\subsection{Objective Metrics}

We evaluate spectral fidelity using Mel-Cepstral Distortion (MCD). Given the ground-truth mel-cepstral coefficients $\mathbf c_t$ and the generated coefficients $\hat{\mathbf c}_{\pi(t)}$, where $\pi(\cdot)$ denotes the frame alignment obtained by dynamic time warping (DTW), MCD is computed as
\begin{equation}
\mathrm{MCD}
=
\frac{1}{T}
\sum_{t=1}^{T}
\frac{10}{\ln 10}
\sqrt{
2
\sum_{m=1}^{M}
\left(
c_{t,m}
-
\hat{c}_{\pi(t),m}
\right)^2
},
\end{equation}
where $T$ is the number of aligned frames and $M$ is the number of mel-cepstral coefficients. Lower MCD indicates better spectral fidelity.

For pitch evaluation, we extract the fundamental frequency using the DIO algorithm in WORLD. Unvoiced frames are excluded from pitch-based evaluation. Let $f_t$ and $\hat{f}_t$ denote the ground-truth and generated $F_0$ values on the aligned voiced frames. The $F_0$ RMSE is defined as
\begin{equation}
F_0\text{-}\mathrm{RMSE}
=
\sqrt{
\frac{1}{|\mathcal V|}
\sum_{t\in\mathcal V}
\left(
f_t-\hat{f}_t
\right)^2
},
\end{equation}
where $\mathcal V$ denotes the set of voiced frames. Lower $F_0$ RMSE indicates more accurate pitch reconstruction.

We also compute the Pearson correlation coefficient between the ground-truth and generated $F_0$ contours:
\begin{equation}
\rho_{F_0}
=
\frac{
\sum_{t\in\mathcal V}
(f_t-\bar f)
(\hat f_t-\bar{\hat f})
}{
\sqrt{
\sum_{t\in\mathcal V}
(f_t-\bar f)^2
}
\sqrt{
\sum_{t\in\mathcal V}
(\hat f_t-\bar{\hat f})^2
}
},
\end{equation}
where $\bar f$ and $\bar{\hat f}$ are the mean ground-truth and generated $F_0$ values over voiced frames. Higher correlation indicates better preservation of pitch contour dynamics.

\subsection{Subjective Evaluation Guidelines}

We conduct MOS evaluation along two dimensions: naturalness and intelligibility. Annotators are asked to evaluate each audio sample independently without knowing which model generated it. The two dimensions are defined as follows.

\paragraph{Naturalness.}
Naturalness measures how human-like and fluent the audio sounds, including smoothness, absence of artifacts, and overall acoustic quality. Annotators are instructed not to focus on whether the content is correct when assigning this score.

\paragraph{Intelligibility.}
Intelligibility measures whether the pronunciation is clear and whether the intended Nüshu syllables can be recognized. Annotators are instructed to focus on linguistic clarity and pronunciation correctness rather than overall audio pleasantness.

For both dimensions, annotators use a five-point scale:
\begin{center}
\footnotesize
\begin{tabular}{cl}
\toprule
\textbf{Score} & \textbf{Description} \\
\midrule
5 & Excellent: very natural or clearly intelligible \\
4 & Good: mostly natural or mostly intelligible \\
3 & Fair: understandable but with noticeable issues \\
2 & Poor: difficult to understand or clearly unnatural \\
1 & Bad: unintelligible or severely degraded \\
\bottomrule
\end{tabular}
\end{center}

\end{document}